\documentclass{article}
\pdfoutput=1

\usepackage{xspace}
\usepackage{times}
\usepackage{epsfig}
\usepackage{graphicx}
\usepackage{amsmath}
\usepackage{amssymb}
\usepackage{algorithm}
\usepackage{algorithmic}
\usepackage{mathtools}
\usepackage{array}
\usepackage{multirow}
\usepackage[inline]{enumitem}
\usepackage{subfig}
\usepackage{multirow}

\usepackage[final,nonatbib]{nips_2017_arxiv}

\usepackage[utf8]{inputenc} 
\usepackage[T1]{fontenc}    
\usepackage{hyperref}       
\usepackage{url}            
\usepackage{booktabs}       
\usepackage{amsfonts}       
\usepackage{nicefrac}       
\usepackage{microtype}      

\usepackage{graphicx}
\usepackage{subfig}

\title{Combining Weakly and Webly Supervised Learning for Classifying Food Images}

\author{
  Parneet Kaur \\
  Rutgers University\\
  New Brunswick, NJ \\
  \texttt{parneet@rutgers.edu} \\
  \And
  Karan Sikka \\
  SRI International\\
  Princeton, NJ \\
  \texttt{karan.sikka@sri.com} \\
   \And
  Ajay Divakaran \\
  SRI International\\
  Princeton, NJ \\
  \texttt{ajay.divakaran@sri.com} \\
}

\begin{document}

\maketitle

\makeatletter
\DeclareRobustCommand\onedot{\futurelet\@let@token\@onedot}
\def\@onedot{\ifx\@let@token.\else.\null\fi\xspace}

\def\etal{et al\onedot}
\def\etc{etc\onedot}
\def\ie{i.e\onedot}
\def\eg{e.g\onedot}
\def\cf{cf\onedot}
\def\vs{vs\onedot}
\def\pd{\partial}
\def\grad{\nabla}
\def\Li{\mathcal{L}}
\def\O{\mathcal{O}}
\def\N{\mathbb{N}}
\def\C{\mathcal{C}}
\def\Lt{\tilde{L}}
\def\R{\mathbb{R}}
\def\X{\mathcal{X}}
\def\I{\mathcal{I}}
\def\F{\mathcal{F}}
\def\w{\textbf{w}}
\def\x{\textbf{x}}
\def\k{\textbf{k}}
\def\k{\textbf{k}}
\def\d{\boldsymbol{\delta}}
\def\y{\textbf{y}}
\def\l{\boldsymbol{\ell}}
\def\wrt{w.r.t\onedot}
\def\a{\boldsymbol{\alpha}}
\def\vertspace{0.6em}


\begin{abstract}
Food classification from images is a fine-grained classification problem. Manual curation of food images is cost, time and scalability prohibitive. On the other hand, web data is available freely but contains noise. In this paper, we address the problem of classifying food images with minimal data curation. We also tackle a key problems with food images from the web where they often have multiple co-occuring food types but are weakly labeled with a single label. We first demonstrate that by sequentially adding a few manually curated samples to a larger uncurated dataset from two web sources, the top-1 classification accuracy increases from $50.3\%$ to $72.8\%$. To tackle the issue of weak labels, we augment the deep model with Weakly Supervised learning (WSL) that results in an increase in performance to $76.2\%$. Finally, we show some qualitative results to provide insights into the performance improvements using the proposed ideas. 
\end{abstract}


\section{Introduction}
\label{sec:intro}
Increasing use of smartphones has generated interest in developing tools for monitoring food intake and trends~\cite{puri2009recognition, zhang2015snap, Mey2015}. Estimate of calorie intake can help users to modify their food habits to maintain a healthy diet.  Current food journaling applications like Fitbit App~\cite{fitbit}, MyFitnessPal~\cite{myfitnesspal} and My Diet Coach~\cite{dietcoach} require users to enter their meal information manually. A study of 141 participants in~\cite{cordeiro2015rethinking} reports that $25\%$ of the participants stopped food journaling because of the effort involved while $16\%$ stopped because they found it to be time consuming. Capturing images of meals is easier, faster and convenient than manual data entry. An automated algorithm for measuring calories from images should be able to solve several sub-problems $-$ classify, segment and estimate 3D volume of the given food items. In this paper we focus on the first task of classification of food items in still images. This is a challenging task due to a large number of food categories, high intra-class variation and low inter-class variation among different food classes. Further, in comparison to standard computer vision problems such as object detection~\cite{lin2014microsoft} and scene classification~\cite{zhou2017places}, present datasets for food classification are limited in both quantity and quality to train deep networks (see~\autoref{sec:relatedwork}). Prior works try to resolve this issue by collecting training data using human annotators or crowd-sourcing platforms~\cite{farinella2016retrieval,chen2012automatic,Kaw2014,zhang2015snap, Mey2015}. Such data curation is expensive and limits the scalability in terms of number of training categories as well as number of training samples per category. Moreover, it is challenging to label images for food classification tasks as they often have co-occurring food items, partially occluded food items, and large variability in scale and viewpoints. Accurate annotation of these images would require bounding boxes, making data curation even more time and cost prohibitive. Thus, it is important to build food datasets with minimal data curation so that they can be scaled for novel categories based on the final application.

\begin{figure}[!t]
	\centering
	\includegraphics[width = 4.5in]{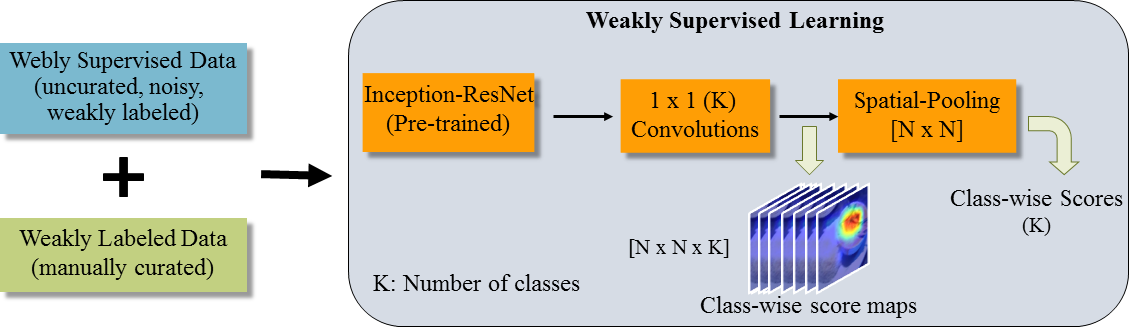}
	\caption{Proposed pipeline for food image classification. We use inexpensive but noisy web data and sequentially add manually curated data to the weakly supervised uncurated data. We also propose to augment the deep model with Weakly Supervised learning (WSL) to tackle the cross-category noise present in web images, and to identify discriminative regions to disambiguate between fine-grained classes.}
	\label{fig:overview}
\end{figure}

\begin{figure}[!t]
	\includegraphics[width = 5.5in]{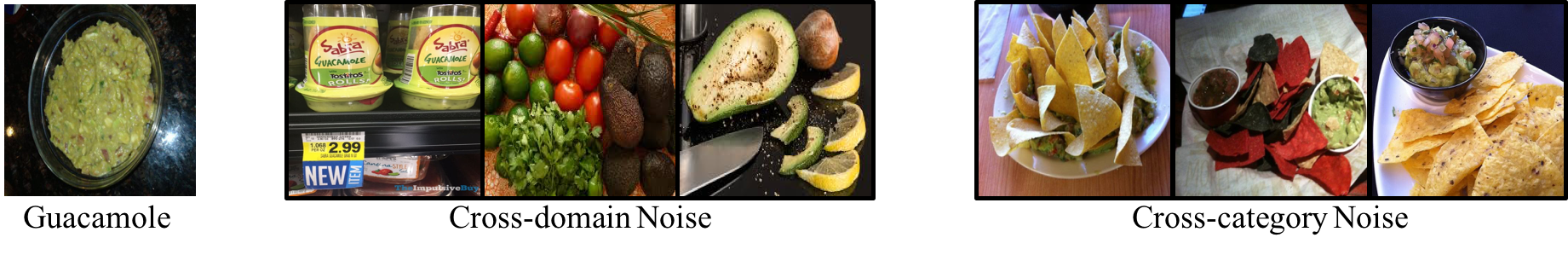}
	\caption{Noise in web data. \emph{Cross-domain Noise}: Along with the images of specific food class, web image search also include images of processed and packaged food items and their ingredients. \emph{Cross-category Noise}: An image may have multiple food items but it has only one label as its ground truth, resulting in cross-domain noise.}
	\label{fig:noise}
\end{figure}

Unlike data obtained by human supervision, web data is freely available in abundance but contains different types of noise~\cite{chen2015webly, wang2008annotating, sukhbaatar2014learning}. Web images collected via search engines may include images of processed and packaged food items as well as ingredients required to prepare the food items as shown in Figure~\ref{fig:noise}. We refer to this noise as cross-domain noise as it is introduced by the bias due to specific search engine and user tags. In addition, the web data may also include images with multiple food items while being labeled for a single food category (cross-category noise). For example, in images labeled as Guacamole, Nachos can be predominant (Figure~\ref{fig:noise}). Further, the web results may also include images not belonging to any particular class.  

We address the problem of food image classification by combining webly and weakly supervised learning (~\autoref{fig:overview}). We first propose to overcome the issues associated with obtaining clean training data for food classification by using inexpensive but noisy web data. In particular we demonstrate that by sequentially adding manually curated data to the uncurated data from web search engines, the classification performance improves linearly. We show that by augmenting a smaller curated dataset with larger uncurated web data the classification accuracy increases from $50.3\%$ to $72.8\%$, which is at par with the performance obtained with the manually curated dataset ($63.3\%$). We also propose to augment the deep model with weakly supervised learning (WSL) for for two reasons (1) tackle the cross-category noise present in web images, and (2) identify discriminative regions to disambiguate between fine-grained classes. We are able to approximately localize food items using the activation maps provided by WSL. We show that by using WSL, the classification accuracy on test data further increases to $76.2\%$. We finally show qualitative results and provide useful insights into the two proposed strategies and discuss the reasons for performance improvements. 



\section{Related Work}
\label{sec:relatedwork}
Traditional computer vision feature vectors such has HOG, SIFT, bag-of-features, Gabor filters and color histograms have been used for classifying food images in~\cite{zhang2015snap, puri2009recognition, Bos2014, bettadapura2015leveraging, joutou2009food}
Recent state-of-the-art deep learning methods for food recognition and localization have led to significant improvement in performance~\cite{liu2016food,Mey2015, liu2016deepfood, Yan2015, Sin2016, Bol2016}. However, the proposed methods use training data with only one food item in the image \cite{liu2016food} or have labels for multiple food items in images~\cite{Kaw2014, Mey2015}. The preparation of training data requires manual curation. The Food-101 dataset~\cite{Bos2014} is often used for food classification. It is collected from a food discovery website \url{foodspotting.com} and generally contains less cross-domain noise as compared to images obtained from search engines such as \url{Google.com}. However, this website relies on images sent by users and thus has limited images for unique food categories, limiting expansion to new categories. In ~\cite{wang2015recipe}, food data is collected from the web but also relies on textual information along with the images.
CNNs have also been used to classify food vs. non-food items in~\cite{Sin2016, Bol2016}. In addition, ~\cite{Bol2016} also provides food activation maps on the input image to generate bounding boxes for localization. We address the problem of classifying food items by using the noisy web data and incorporating weakly supervised learning for training CNNs. 

Recent approaches of webly supervised learning in computer vision leverage from the noisy web data, which is easy and inexpensive to collect. Prior work uses web data to train CNNs for classification and object detection. \cite{Kra2016} use noisy data collected from web for fine-grained classification. They also use active learning-based approach for collecting data when only limited examples are available from web. They demonstrate that even if the classification task at hand has small number of categories, using a network trained with more categories gives better performance. Motivated by curriculum learning, \cite{chen2015webly}, propose an algorithm to first train a model on simple images from Google and estimate a relationship graph between different classes. The confusion matrix is integrated with the model and is fine-tuned on harder Flickr images. The confusion matrix makes the network robust to noise and improves performance. Similarly, ~\cite{patrini2016making} modified the loss function by using the noise distribution from the noisy images.  

Food images often consist of multiple food items instead of a single food item and require bounding boxes for annotation. To avoid expensive curation, weakly supervised learning (WSL) utilizes image-level labels instead of pixel-level labels or bounding boxes. In \cite{Zhou_2016_CVPR, oquab2015object}, the network architecture is modified to incorporate WSL by adding a global pooling layer. Along with image classification, these architectures are able to localize the discriminative image parts. In \cite{durand2016weldon}, the authors include top instances (most informative regions) and negative evidences(least informative regions) in the network architecture to identify discriminative image parts more accurately. To address object detection, the authors in ~\cite{bilen2016weakly} modify the deep network using a spatial pyramid pooling layer and use region proposals to simultaneously select discriminative regions and perform classification. In ~\cite{cinbis2017weakly}, the authors present  a multi-fold multiple instance learning approach that detects object regions using CNN and fisher vector features while avoiding convergence to local optima.

In this paper, we combine the webly and weakly supervised learning to address the problem of food classification. We sequentially add curated data to the weakly labeled uncurated web data and augment the deep model with WSL. We report improved performance as well as gain insights by visualizing the qualitative results.

\section{Approach}
\label{sec:approach}
We first describe the datasets used to highlight the benefits of using uncurated data with manually curated data for the task of food classification. Thereafter, we briefly discuss weakly supervised learning to train the deep network. 

\subsection{Datasets}
We first collect food images from the web and augmented it with both curated and additional uncurated images, and test our method on a separate clean test set. The datasets are described below:

\begin{enumerate}
\item \textbf{Food-101}~\cite{Bos2014}: This dataset consists of $101$ food categories with $750$ training and $250$ test images per category. The test data was manually cleaned by the authors whereas the training data consists of cross-category noise \ie images with multiple food items labeled with a single class. We use the manually cleaned test data as the curated dataset ($25k$ images), \textit{Food-101-CUR}, which is used to augment the web dataset. We use $10\%$ of the uncurated training data for validation and $90\%$ of uncurated data (referred to as \textit{Food-101-UNCUR}) for data augmentation for training the deep model. 

\item \textbf{Food-Web-G}: We collect the web data using Google image search for food categories from Food-101 dataset~\cite{Bos2014}. The restrictions on public search results limited the collected data to approximately $800$ images per category. We removed images smaller than $256$ pixels in height or width from the dataset. As previously described, the web data is weakly labeled and consists of both cross-domain and cross-category noise as shown in ~\autoref{fig:noise}. We refer to this dataset as \textit{Food-Web-G}

\item \textbf{UEC256}~\cite{Kaw2014}:
This dataset consists of $256$ food categories, including Japanese and international dishes and each category has at least $100$ images with bounding box indicating the location of its category label. Since this dataset provides the advantage of complete bounding box level annotations, we use this dataset for testing. We construct the test set by selecting $25$ categories in common with the Food-101 dataset and extract cropped images using the given bounding boxes. 
\end{enumerate}  

\subsection{Weakly Supervised Learning (WSL)}
The data collected from web using food label tags is weakly labeled \ie an image is labeled with a single label when it contains multiple food objects. We observe that most uncurated food images were unsegmented with images containing either items from co-occurring food classes or background objects such as kitchenware. We propose to tackle this problem by augmenting the deep network with WSL that explicitly grounds the discriminative parts of an image for the given training label \cite{Zhou_2016_CVPR}, resulting in a better model for classification. 

As shown in \autoref{fig:overview}, we incorporate discriminative localization capabilities into the deep model by adding a $1 \times 1$ convolution layer and a spatial pooling layer to a pretrained CNN \cite{oquab2015object, Zhou_2016_CVPR}. The convolution layer generates $N \times N \times K$ class-wise score maps from previous activations. The spatial pooling layer in our architecture is a global average pooling layer which has recently been shown to outperform the global max pooling step for localization in WSL~\cite{oquab2015object, Zhou_2016_CVPR}. Max pooling only identifies the most discriminative region by ignoring lower activations, while average pooling finds the extent of the object by recognizing all discriminative regions and thus giving better localization. The spatial pooling layer returns class-wise score for each image which are then used to compute cross-entropy loss.  
During test phase, we visualize the heats maps for different classes by overlaying the predicted score maps on the original image. 

Additionally, food classification is a fine-grained classification problem \cite{Kra2016} and we later show that discriminative localization also aids in correctly classifying visually similar classes. Compared to Krause \etal \cite{Kra2016}, who show the benefits of noisy data for fine-grained tasks such as bird classification, we also highlight the benefits of WSL for learning with noisy data for food classification. 

\subsection{Implementation Details}
We use Inception-Resnet~\cite{szegedy2017inception} as the base architecture and fine-tune the weights of a pre-trained network (ImageNet). During training, we use Adam optimizer with a learning rate $10^{-3}$ for the last fully-connected (classification) layer and $10^{-4}$ for the pre-trained layers. We use batch size of $50$. For WSL, we initialize the network with the weights obtained by training the base model and only fine-tune the layers added for weak localization with learning rate of $ 10^{-3}$. For WSL we obtain localized score maps for different classes by adding a $1 \times 1$ convolutional layer to map the input feature maps into classification score maps \cite{Zhou_2016_CVPR}. For an input image of $299 \times 299$, we get a score map of $8 \times 8$ from the output of this convolutional layer, which gives approximate localization when resized to the size of input image. The average pooling layer is of size $8 \times 8$, stride $1$ and padding $1$.


\section{Experiments}
\label{sec:experiments}

\subsection{Quantitative Results}

\def\arraystretch{1}
\begin{table*}[t]
	\centering
	\newcolumntype{C}{>{\centering\arraybackslash}p{5em}}
	\begin{tabular}{|C|C|C|C|C|C|C|C|C|}
		\hline
		\multicolumn{5}{|c|}{Dataset} & No. of images & Type & w/o WSL & with WSL  \\ \hline
		\multicolumn{5}{|c|}{Food-Web-G} & $66.9k$ & $N$ & $55.3\%$ & $61.6\%$ \\ \hline
		\multicolumn{5}{|c|}{Food-101-CUR} & $25k$ & $C$ & $63.3\%$ & $64.0\%$ \\ \hline
		\multicolumn{5}{|c|}{Food-101-UNCUR} & $67.5k$ & $N$ & $70.5\%$ & $73.2\%$ \\ \hline
		\multicolumn{5}{|c|}{Food-Web-G + Food-101-CUR} & $92.5k$ & $N+C$ & $69.7\%$ & $73.0\%$ \\ \hline
		\multicolumn{5}{|c|}{Food-Web-G + Food-101-UNCUR} & $134.4k$ & $N$ & $70.1\%$ & $74.0\%$ \\ \hline
		\multicolumn{5}{|c|}{Food-Web-UNCUR + Food-101-CUR} & $92.5k$ & $N+C$ & $71.4\%$ & $75.1\%$ \\ \hline
		\multicolumn{5}{|c|}{All datasets} & $159.3k$ & $N+C$ & $\mathbf{72.8}\%$ & $\mathbf{76.2}\%$ \\ \hline
	\end{tabular}
	\caption{Classification accuracy for different combinations of datasets with and without Weakly Supervised training. The number of images for each combination ($k=1000$) and type of dataset ($N$: noisy and $C$:clean) are also shown.}
	\label{tbl:tbl_classification_accuracy}
\end{table*}

\begin{figure}[!ht]
	\centering
	\subfloat[]{
		\includegraphics[width = 2.7in]{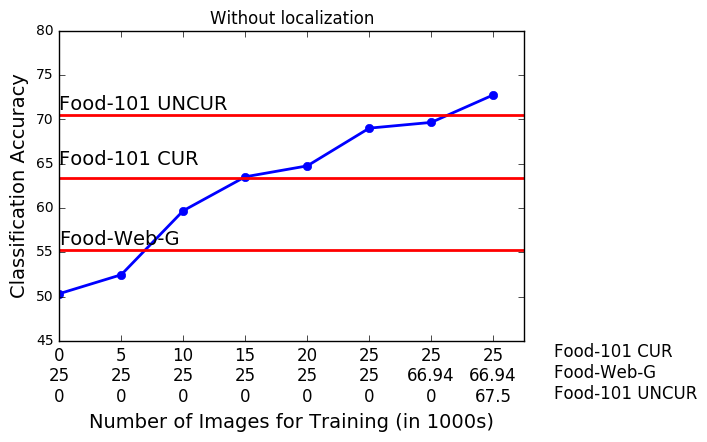}} 
	\subfloat[]{
		\includegraphics[width = 2.7in]{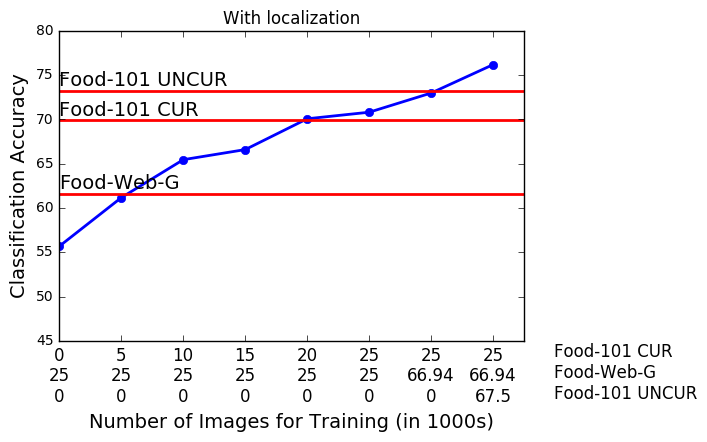}}
	\caption{Classification Accuracy using (a) Inception Resnet, (b) Inception Resnet with localization layer. As the curated data (Food-101 CUR) is added to the web data, the classification accuracy on the test data (UEC256-test) increases. Increasing the web data results in further improvement. The red line shows the baseline performance with individual datasets.}
	\label{fig:classification_accuracy}
\end{figure}

We report top-1 classification accuracy for different combinations of datasets (see \autoref{sec:approach}) and WSL in ~\autoref{tbl:tbl_classification_accuracy}. We first discuss the performance without WSL, where the baseline performance using Google images (Food-Web-G) is $55.3\%$. We observe that augmenting Food-Web-G ($66.9k$ samples) with a small proportion of curated data ($25k$ samples) improves the performance to $69.7\%$, whereas augmentation with additional uncurated data ($67.5k$ samples from \url{foodspotting.com}) results in $70.1\%$. The performance of both combinations is higher compared to the curated data alone ($63.3\%$) clearly highlighting the performance benefits of using noisy web data. We also observe that different sources of web images \ie Google versus Foodspotting results in different performance ($55.3\%$  versus $70.5\%$ respectively) for similar number of training samples. As previously mentioned, Foodspotting is crowdsourced by food enthusiasts, who often compete for ratings, and thus has less cross-domain noise and better quality compared to Google images. By combining all the three datasets, we observe a classification accuracy of $72.8\%$, which outperforms the performance obtained by either curated and uncurated datasets alone.

We also wanted to study the variation in performance on using different proportions of clean and unclean images. As shown in ~\autoref{fig:classification_accuracy}, by sequentially adding manually curated data (Food-101 CUR) to the web data (Food-Web-G), the classification performance improves linearly from $50.3\%$ to $69.0\%$. By adding the uncurated data from foodspotting, it further increases to $72.8\%$. We also observe significant improvements by adding discriminative localization to the deep model, where the classification accuracy further increases to $76.2\%$. In particular we observe a consistent improvement across all data splits by using WSL \eg for the combination of both uncurated datasets from Google and foodspotting, the hike in performance by using WSL is $4\%$ absolute points. This performance trend highlights the advantages of WSL in tackling the noise present in food images by implicitly performing foreground segmentation and also focusing on the correct food item in the case when multiple food items are present (cross-category noise).

\subsection{Qualitative Results}

\begin{figure}[!t]
	\centering
	\subfloat[]{
		\includegraphics[height = 1.5in]{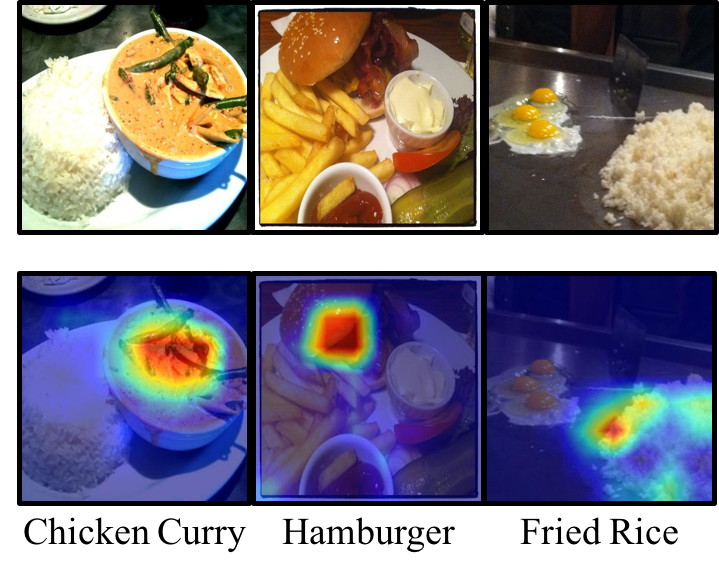}
		\label{fig:heatmaps_train_1}		
		} 
	\subfloat[]{
		\includegraphics[height = 1.5in]{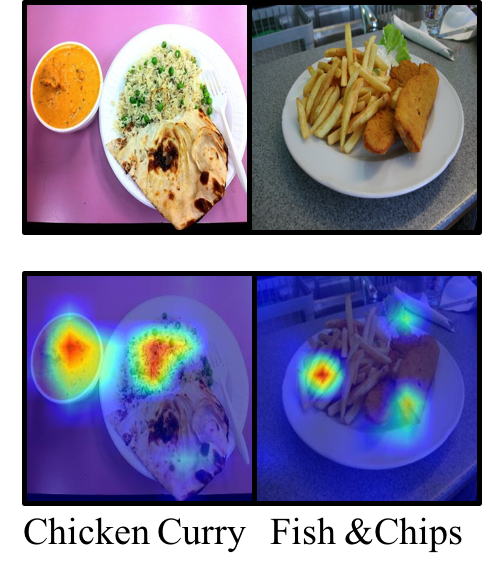}
		\label{fig:heatmaps_train_2}			
		}\quad
	\subfloat[]{
		\includegraphics[height = 1.5in]{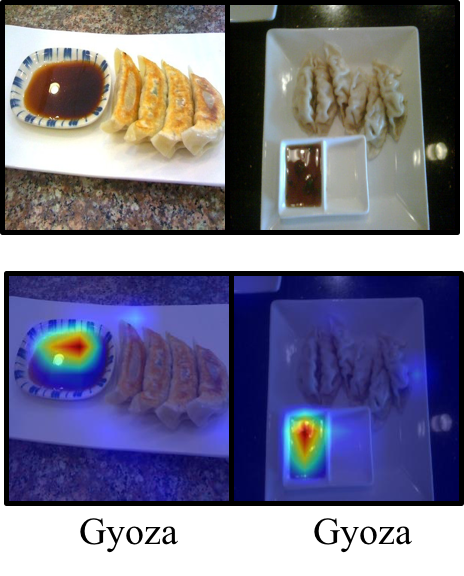}
		\label{fig:heatmaps_train_3}			
		}	
	\caption{Heat maps showing approximate pixel-wise predicted probabilities obtained by Weakly Supervised training for few training images. We show three cases where (a) the food items are localized correctly, (b) network localizes frequently co-occurring food items due to weak labels for training, and (c) network localizes a frequently co-occurring food item instead of the labeled food item due to incomplete and noisy training data.}
	\label{fig:heatmaps_train}
\end{figure}

\begin{figure}[!t]
	\centering
	\includegraphics[width = 4.3in]{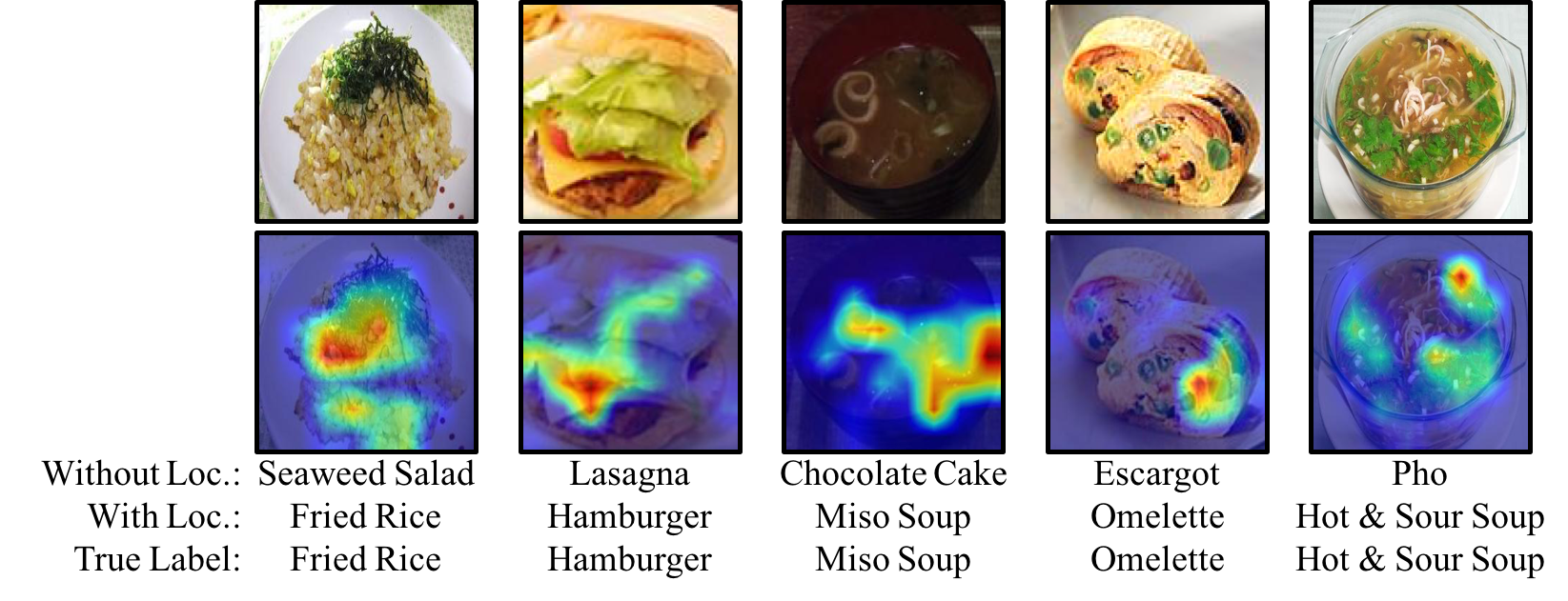}
	\caption{Test images that are misclassified without any localization but correctly classified with weak localization. We also show the heat map, predicted label with the two approaches, and the true label.}
	\label{fig:heatmaps_test}
\end{figure}

We show the heat maps indicating the approximate localization of the top-1 predicted label for few training images with multiple food items in~\autoref{fig:heatmaps_train}. We see that for some training images (Figure \autoref{fig:heatmaps_train_1}) the network learns to correctly localize the correct food type among co-occurring food classes \eg it is able to identify rice in ``fried rice'' example. This ability could explain the reasons for performance benefits especially when training data is not completely labeled. However, we also observe that for frequently co-occurring food items, sometimes the network learns to localize multiple food types together. As shown in Figure \autoref{fig:heatmaps_train_2}), network learns ``chicken'' and ``rice'' as one category because they co-occur in many training examples. The network also learns wrong food item for some co-occurring food items. For example, Figure \autoref{fig:heatmaps_train_3} shows some examples where the network learns to recognize ``sauce'' instead of Gyoza. This is a drawback with standard WSL methods where the algorithm generally tends to focus on the most discriminative part and overfits. We can overcome this aspect by either leveraging additional clean training data or using recent advances in WSL \cite{Singh_2017_ICCV, Kim_2017_ICCV}. We show the heat maps for test images that are misclassified without localization but are correctly classified with localization in~\autoref{fig:heatmaps_test}. Food classification is a fine-grained classification problem and we can see that WSL helps by identifying discriminative parts for different food items. For \eg, the model grounds the noodle pieces in ``miso soup'' image in \autoref{fig:heatmaps_test} that makes it possible to differentiate it from ``chocolate cake'' class, both of which are generally dark brown in color.    

\begin{figure}[!t]
	\centering
	\subfloat[]{
		\includegraphics[height = 2in]{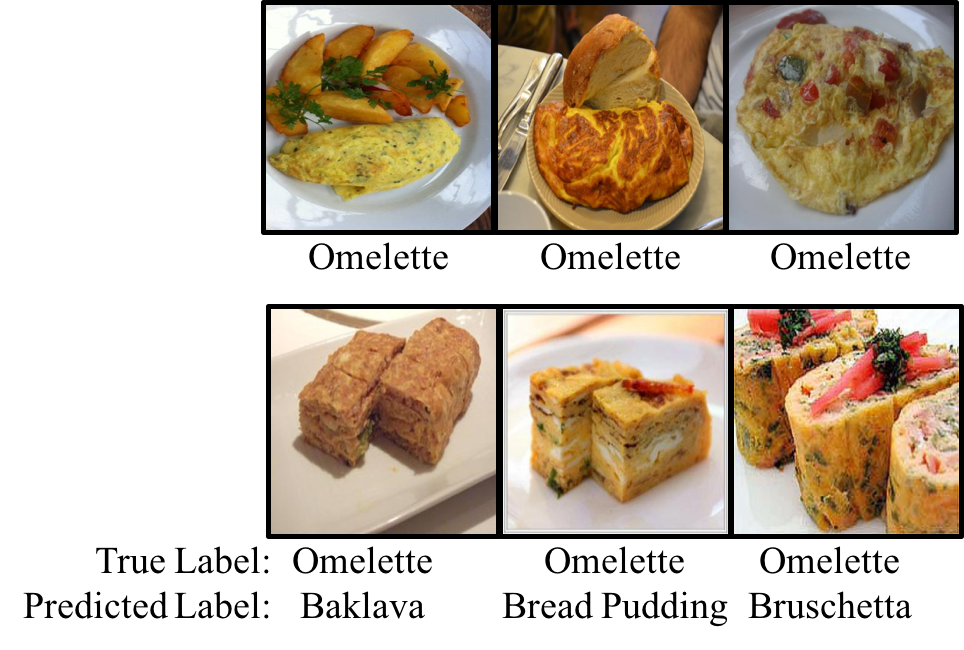}
		\label{fig:misclassification_1_a}
	} \quad
	\subfloat[]{
		\includegraphics[height = 		2in]{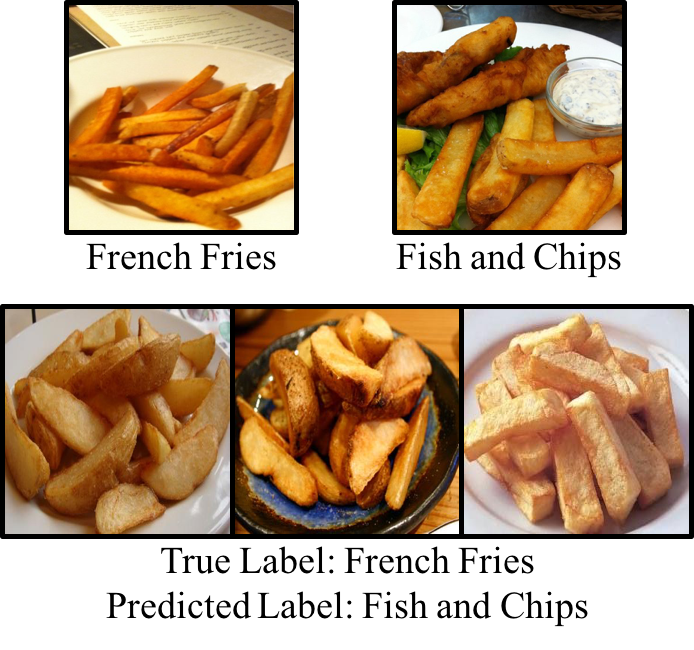}		\label{fig:misclassification_1_b}
	}
	\caption{Misclassification in test data. (a) Examples for category `Omelette' in training data (top row) and test data (bottom row). Data distribution in training and test data differs because they are collected from different sources, resulting in  misclassification of some test images. (b) Examples from training data (top row) and test data (bottom row). Inter-class similarity in training data causes confusion and results in misclassification of test data.}
	\label{fig:misclassification_1}
\end{figure}

We observe that the properties of training data and quality of labeling influences the test performance. 
There are unique ways of cooking a food item in different cuisines, resulting in variability in appearance. The UEC256 test data mainly contains Japanese cuisines that may not be seen by the network during training phase. We found that some test images are misclassified if their appearance varies from the training images. Figure~\autoref{fig:misclassification_1_a} shows an example of the category ``omelette'' that has high variability for training and test data.
We observe that the performance on test data is also influenced by the weak/incomplete labeling of training data. For example, as shown in Figure~\autoref{fig:misclassification_1_b}, the training dataset contains these two categories: ``french fries'' and ``fish and chips''. ``Fish and chips'' always contains \textit{french fries}, however this information is not used during the training phase resulting in high confusion between these classes during testing. 

\begin{figure}[!t]
	\centering
	\includegraphics[width = 5in]{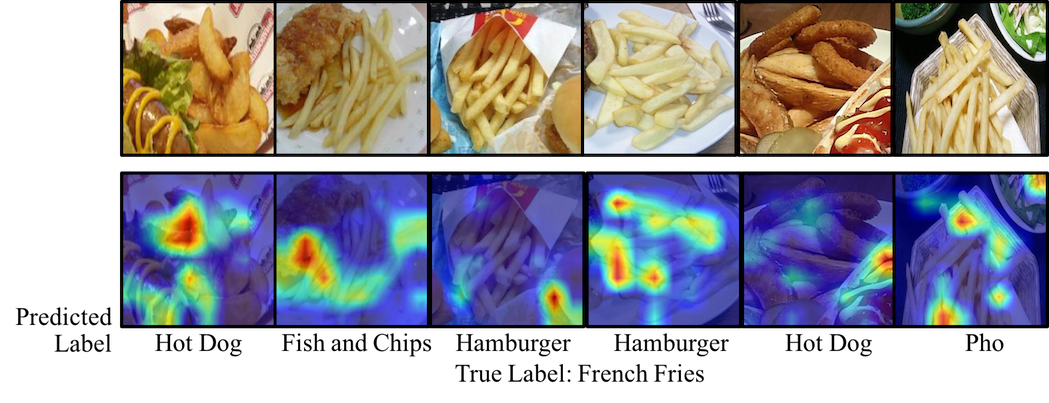}
	\caption{The test images are misclassified in presence of occluded food items because the network learns to localize co-occurring food items during training. For these images, the top-5 predicted labels includes the ground truth.}
	\label{fig:misclassification_2}
\end{figure}

Misclassification on test images also occurs due to the presence of multiple food items. Localization heatmaps show that the network also focuses on the partially occluded food items in the images. \autoref{fig:misclassification_2} shows some examples where the test images with true label ``french fries'' are misclassified because the network focuses on the other partial food items in the image. Even though the top-most predicted label corresponds to the partially occluded food item, the correct label is often found in the top-5 predictions (top-5 accuracy is $90.8\%$).
 
\begin{figure}[!t]
	\centering
	\includegraphics[width = 5.5in]{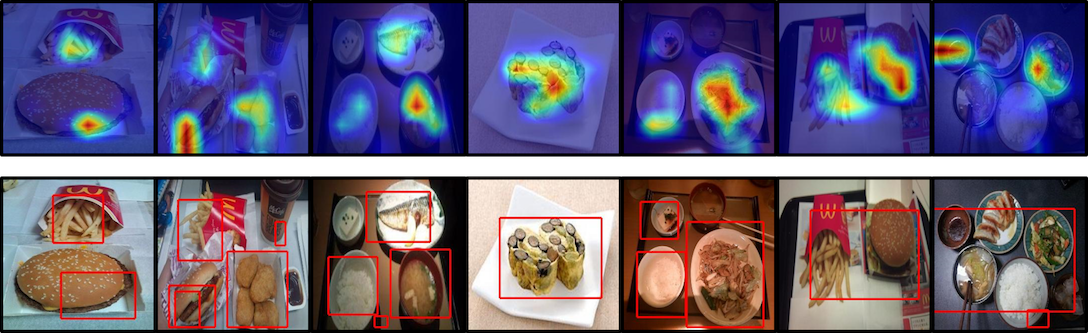}
	\caption{ UEC256 test images. [Top row] Heatmaps for the top-1 predicted class. [Bottom row] Bounding boxes obtained using the heatmaps.}
	\label{fig:heatmaps_test_full}
\end{figure}

We also generated bounding boxes from the heatmaps as shown in ~\autoref{fig:heatmaps_test_full} and will evaluate the localization performance in the future.

\section{Conclusion}
\label{sec:conclusion}
In this paper, we leverage the freely available web data to address the problem of food classification. 
By augmenting the abundantly available uncurated web data with limited manually curated dataset and using weakly supervised learning, we achieve a classification accuracy of $76.2\%$. The performance improves linearly as the amount of curated data for training is increased. We examine the localization maps and observe that WSL aids the network by learning to approximately localize a food item even in presence of multiple food items. Additionally, we examine some cases where discriminative localization helps to disambiguate visually similar classes. Although we chose to focus on WSL in this work, additional performance improvement can also be obtained by other complementary approaches such as cost sensitive loss \cite{chen2015webly, patrini2016making} and domain adaptation \cite{bergamo2010exploiting}.

\section{Acknowledgments}
We thank Carter Brown, Ankan Bansal, Kilho Son and Anirban Roy for many helpful discussions.

\end{document}